\pdfoutput=1

\documentclass[11pt]{article}

\usepackage[final]{acl}
\usepackage{natbib}
\usepackage{times}
\usepackage{latexsym}
\usepackage{multirow}
\usepackage{siunitx}
\usepackage{amsmath}
\usepackage{amssymb}
\usepackage{booktabs}

\usepackage[T1]{fontenc}

\usepackage[utf8]{inputenc}

\usepackage{microtype}

\usepackage{inconsolata}

\usepackage{graphicx}
\usepackage{booktabs}
\title{RAG-RLRC-LaySum at BioLaySumm: Integrating Retrieval-Augmented Generation and Readability Control for Layman Summarization of Biomedical Texts}
\author{
 \textbf{Yuelyu Ji\textsuperscript{1}},
 \textbf{Zhuochun Li\textsuperscript{1}},
  \textbf{Rui Meng\textsuperscript{2}}, 
  \textbf{Sonish Sivarajkumar\textsuperscript{1}},
\\
 \textbf{Yanshan Wang\textsuperscript{1}},
 \textbf{Zeshui Yu\textsuperscript{1}},
 \textbf{Hui Ji\textsuperscript{1}},
  \textbf{Yushui Han\textsuperscript{1}},
 \\
 \textbf{Hanyu Zeng \textsuperscript{1}},
  \textbf{Daqing He\textsuperscript{1}}
\\
\\
 \textsuperscript{1}University of Pittsburgh,
 \textsuperscript{2}Salesforce Research
\\
}

\begin{document}

\maketitle
\begin{abstract}
This paper introduces the RAG-RLRC-LaySum framework, designed to make complex biomedical research understandable to laymen through advanced Natural Language Processing (NLP) techniques. Our Retrieval Augmented Generation (RAG) solution, enhanced by a reranking method, utilizes multiple knowledge sources to ensure the precision and pertinence of lay summaries. Additionally, our Reinforcement Learning for Readability Control (RLRC) strategy improves readability, making scientific content comprehensible to non-specialists. Evaluations using the publicly accessible PLOS and eLife datasets show that our methods surpass Plain Gemini model, demonstrating a 20\% increase in readability scores, a 15\% improvement in ROUGE-2 relevance scores, and a 10\% enhancement in factual accuracy. The RAG-RLRC-LaySum framework effectively democratizes scientific knowledge, enhancing public engagement with biomedical discoveries \footnote{Our code and implementation details are available here: https://github.com/JoyDajunSpaceCraft/RAG-RLRC-LaySum}.


\end{abstract}

\section{Introduction}

Biomedical research encompasses crucial discoveries, ranging from everyday health concerns to significant disease outbreaks. Such studies are essential not only for scientists and doctors but also for journalists and the general public. However, the specialized and complex language typical in these studies often renders the content incomprehensible to those without a scientific background \cite{thoppilan2022lamda}.
\begin{figure}[t]
  \includegraphics[width=\columnwidth]{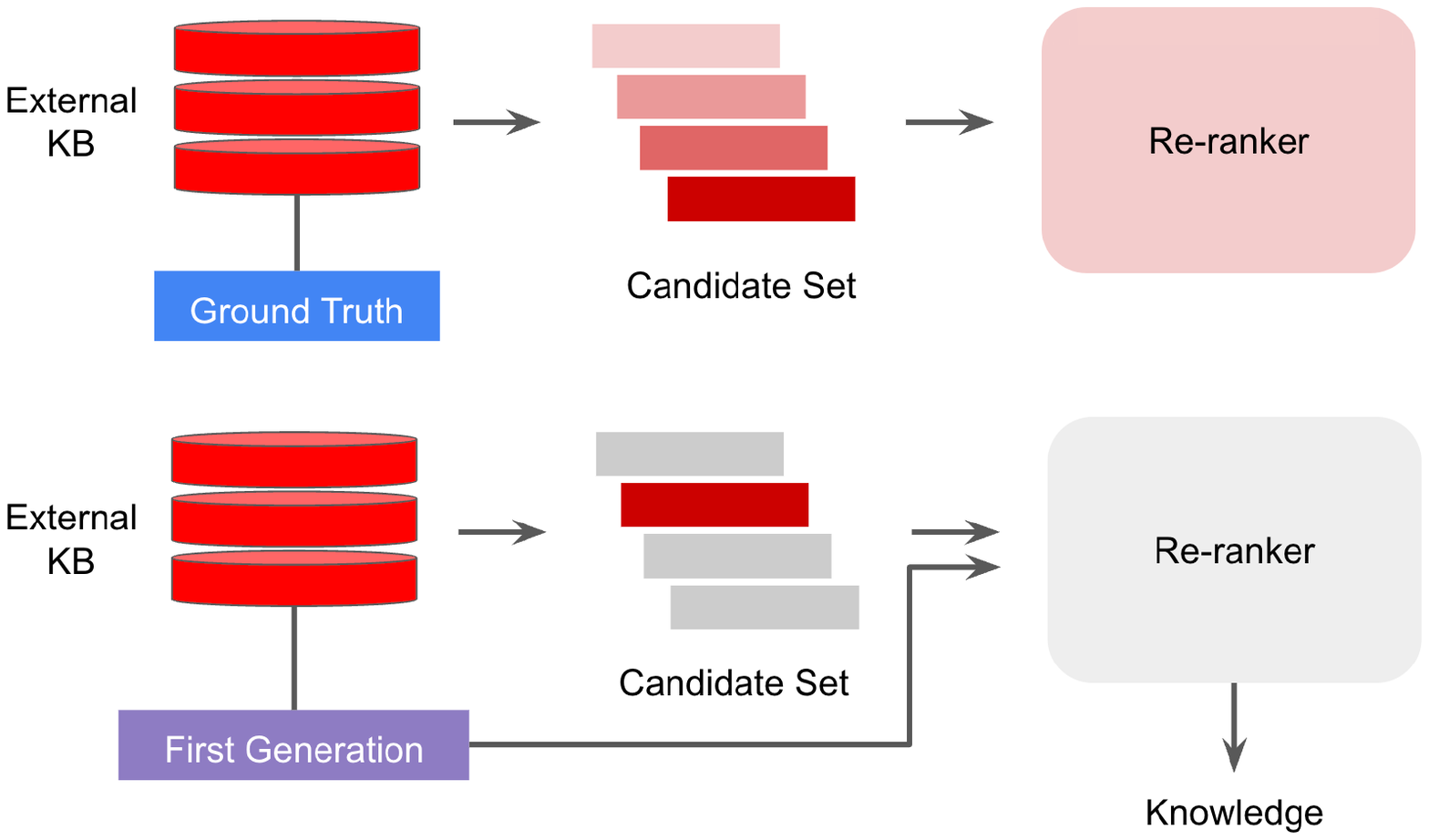}
  \caption{Knowledge Retrieval Augmented, with the trained re-ranker, can provide more relevant knowledge based on the first generation. }
  \label{fig:experiments}
\end{figure}
To address this issue, the development of automated lay summaries have become increasingly important \cite{goldsack-etal-2023-enhancing,goldsack-etal-2023-biolaysumm}. This initiative aims to summarize the detailed aspects of biomedical research into summaries that are both comprehensible and devoid of complicated jargon. Although these systems show great potential, doubts about their accuracy are a major obstacle to their widespread use\cite{gabriel2020go,maynez2020faithfulness,yang-24-data-aug,li-23-deception-detection,li-24-vqa,wang2024infuserki,Zhao_Li_Hong_Zhu_Liu_Dai_2024,zhu2024cross,zhu2024exploiting,li2024utilizing}. Our framework integrates specific external explanations for complex terms to further enhance content simplification.  In response to the concerns about the integrity of summarized information, our framework employs a ``knowledge retrieval'' approach within the Retrieval-Augmented Generation (RAG) framework. This method uses a neural re-ranker to dynamically integrate trustworthy external knowledge sources like Wikipedia, ensuring that summaries are simplified, factually accurate, and contextually relevant\cite{lewis2020retrieval,kang2024knowledge}. 

The architecture of the proposed RAG is illustrated in Figure \ref{fig:experiments}.
\begin{figure*}[t]
  \includegraphics[width=16cm]{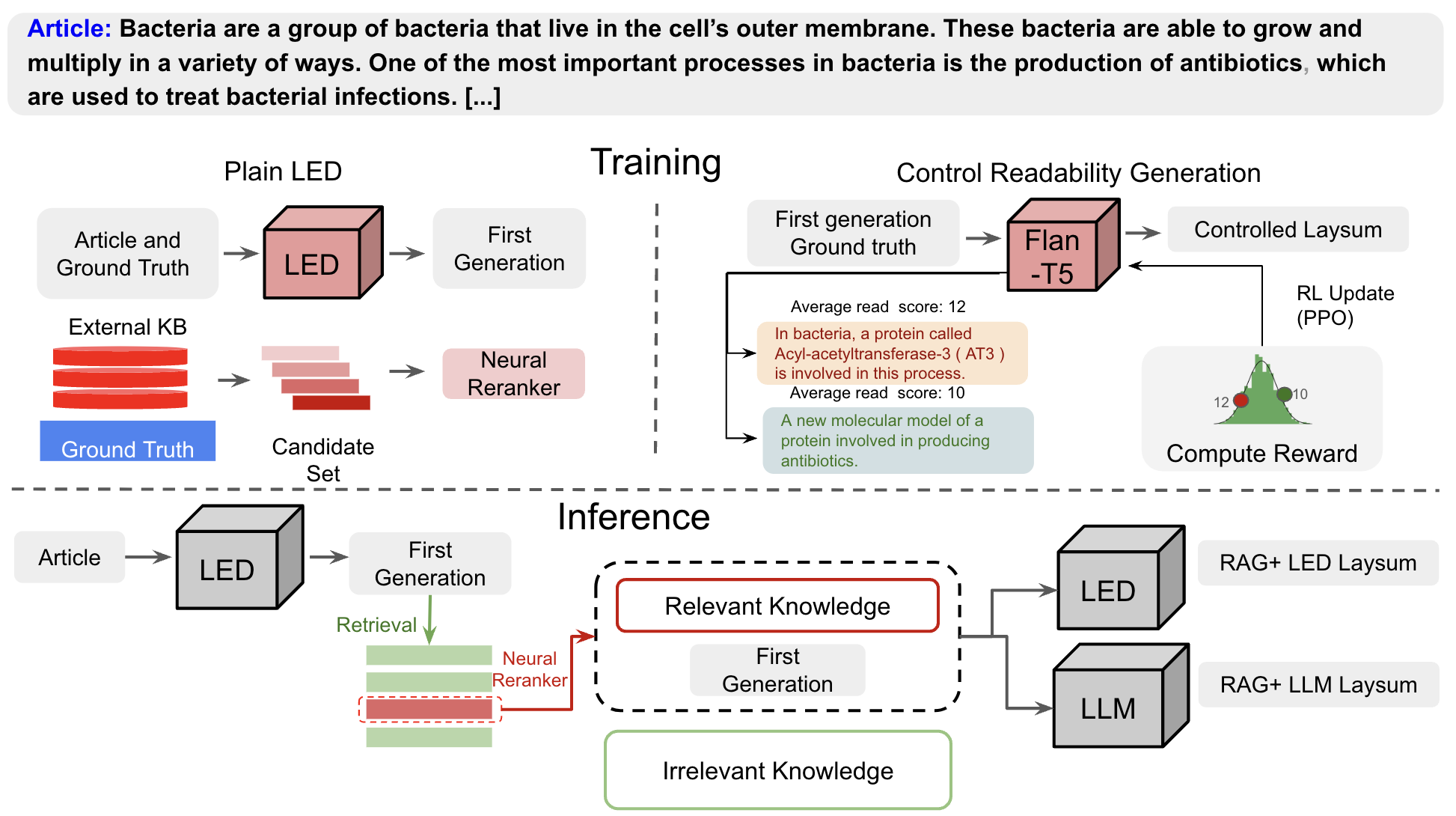}
  \caption{This figure illustrates the architecture of the proposed RAG-RLRC-LaySum model. During the training phase, we employ the  Longformer Encoder-Decoder (LED) model as the backbone~\cite{beltagy2020longformer}. We enhance the model's capabilities through Wikipedia knowledge retrieval during inference. We utilize large language models (LLMs) such as ChatGPT and Gemini to further improve readability and enhance textual clarity by modifying prompts. For controlled text generation, readability scores are utilized to guide the model in generating expected outputs. The outputs of these scores are normalized to ensure text consistency and quality across generated texts.}
  \label{fig:result}
\end{figure*}
We also introduce a reward-based approach to overcome the limitations of traditional fine-tuning, which often produces summaries that have high ROUGE scores but are not actually readable to humans. This method fine-tunes the model by rewarding outputs that align with readability metrics (Flesch-Kincaid Grade Level and Dale-Chall Readability Score \cite{foster2002readability, Ribeiro2023GeneratingSW}). Unlike traditional supervised methods that might limit the model`s adaptability, our approach encourages the model to alternative expressions to enhance clarity. 
The BioLaySumm challenge is a research competition that focused on developing and benchmarking models for generating lay summaries from complex biomedical literature\footnote{https://biolaysumm.org/}. Our method is ranked 11th on the leaderboard of the challenge~\cite{goldsack-etal-2024-biolaysumm}.

\section{Methodology}
First, we employ a Retrieval-Augmented Generation (RAG) solution that processes entire papers despite limited input capacity. Second, we improve summary quality by optimizing readability using relevant background information. The framework is illustrated in Figure \ref{fig:result}.
\subsection{Retrieval Augmented Generation (RAG)}
Our RAG framework enhances keyword-based retrieval by using an initial lay summary generated by the model as a query. In the inference stage, it retrieves relevant descriptions from Wikipedia \cite{ponzetto2007api} by the Pyserini \cite{lin2021pyserini} index. However, retrieving relevant information from a large number of articles remains a challenge because the first generated summaries cannot work as effective queries. We initially use the ground truth as a query but switch to the first generated layman summary during inference for passage retrieval. However, there's a risk that the top-k passages may not be the most relevant for generating accurate summaries.
Given a scientific document $D$ with a set of candidate passages $K=\{k_1,k_2,...,k_n\}$ from grounding sources, the RAG framework generates a lay summary $S$ by maximizing the probability:

\begin{equation}
p(S|D, K) = \prod_{i=1}^{|S|} p(s_i | s_{<i}, D, K)
\end{equation}

where $s_i$ represents the $i$-th token in the summary, and $s_{<i}$ denotes the sequence of tokens preceding $s_i$. We use ColBERT \cite{khattab2020colbert} and BGE-v2 \cite{li2023making,chen2024bge} as two different types of the neural re-ranker. The details about the trained re-ranker are in Appendix ~\ref{sec:Retrieval Design}.
\subsection{Reinforcement Learning for Readability Control (RLRC)}
 For details about the reranking model and sequence generation model training can be seen in Appendix ~\ref{sec:Finetuning model}.
The RLRC method inputs the first generation from the plain LED and uses the ground truth as the expected output. Our RLRC approach employs a reinforcement learning strategy to fine-tune the readability of summaries. We define a reward function  $R(y, r^*)$ based on the desired readability level $r^*$ that encourages the generation of text towards better readability, which is measured by the Flesch Reading Ease score $r^*$:

\begin{equation}
\small
R(y, r^*) = 1 - \exp\left(-\frac{(R(y) - r^*)^2}{2\sigma^2}\right)
\end{equation}

where $R(y)$ denotes the readability score of the generated summary $y$, and $\sigma$ is a hyperparameter that controls the sensitivity of the reward function to deviations from the target readability score. Also, we leverage a Gaussian-based reward that strongly penalizes great variations in the readability \cite{Ribeiro2023GeneratingSW}.

We employ the Proximal Policy Optimization (PPO) \cite{schulman2017proximal} algorithm to optimize our RLRC model. The objective is to adjust the model's parameters $\theta$ by maximizing the objective function:

\begin{equation}
\small
L(\theta) = \mathbb{E}_{(y, r^*) \sim p_{\theta_{\text{old}}}} \left[ \left(\frac{p_{\theta}(y \mid D, r^*)}{p_{\theta_{\text{old}}}(y \mid D, r^*)}\right) R(y, r^*) \right]
\end{equation}

Here, $p_{\theta_{old}}$and $p_{\theta}$ denote the policy under the old and current parameters, respectively.
\subsection{Large Language Models}
We use the LLMs in two ways: first, as a paraphrasing tool during inference to refine initial generations, and second, for directly generating layman summaries. This implementation is built on Gemini-1.0-pro, developed by Google \cite{team2023gemini}, which also serves as our baseline LLM. We aim to create readable summaries while incorporating as many input keywords as possible. We follow Gemini-1.0-pro's default settings, and the prompt details are described in Appendix ~\ref{sec:Gemini Prompt}.

\begin{table*}[htbp]
\caption{Results on PLOS and eLife validation datasets. For the ↑ means, the higher, the better; for the ↓ means, the lower, the better. All best results are marked as bold.  The RAG+different models represent the models that used neural re-ranker. }
\begin{center}
\small 
\setlength\tabcolsep{1pt}
\label{tab:results}
\begin{tabular}{c|c|c|c|c|c|c|c|c|c|c}
\toprule
\multirow{2}{*}{Method}& \multicolumn{4}{c|}{Relevance} & \multicolumn{4}{c|}{Readability} & \multicolumn{2}{c}{Factuality } \\
\cline{2-11}
 & {Rouge1\(\uparrow\) } & {Rouge2\(\uparrow\) } & {RougeL\(\uparrow\) } & {BERTScore\(\uparrow\) } & {FKGL\(\downarrow\)} & {DCRS\(\downarrow\)} & {CLI\(\downarrow\)}&{LENS\(\uparrow\) } & {AlignScore\(\uparrow\) } & {SummaC\(\uparrow\) } \\
\midrule
\multicolumn{11}{c}{PLOS} \\
\hline

Plain LED & 45.96    & \textbf{15.00}    & 41.33    & \textbf{85.97}      & 15.17  & 12.26  & 16.42 &54.96&\textbf{81.68}       & \textbf{74.34}     \\
Plain retrieve+LED & 45.53 & 14.37 & 41.10 & 85.66 & 15.06 & 12.10 & 16.32 & 51.82& 77.38 & 71.94 \\
RAG+LED & 45.64 & 15.37 & \textbf{42.30} & 85.21 & 15.22 & 11.93 &15.92 & 53.42 & 76.57 & 72.83 \\

RAG+ChatGPT & 37.39&6.81&33.96&84.70&\textbf{11.21}&\textbf{10.37}&\textbf{12.50} &71.90 &65.71 &57.85\\ 
RAG+Gemini &38.89&8.74&35.11&85.12&11.33&10.48&13.38&\textbf{74.76} &68.40&58.88\\ 
Plain Gemini & 44.67  & 13.36    & 40.26    & 85.87       & 15.71  & 11.84  & 17.98 &62.64 & 74.18       & 52.82    \\
RAG-RLRC & \textbf{46.58} & 14.96 & 41.81 & 85.83 & 14.89 & 11.81 & 16.78 & 47.55 & 78.45 & 72.97  \\
\midrule
\multicolumn{11}{c}{eLife} \\
\midrule
Plain LED   & 47.02    & 12.52    & 44.07    & \textbf{84.73}      & 10.52  & 9.33   & 11.49& 73.45& \textbf{62.37}       & 60.12  \\
Plain retrieve+LED &  47.72& 12.40 & 44.26& 84.41 & 12.11 & 9.25 & 11.40 & 67.57& 53.57 & 56.18 \\
RAG+LED &  47.69& 12.41& 44.34 & 84.41 & 11.99 & 9.25 & \textbf{11.39} &  67.95 & 53.89 & 55.65\\
RAG+ChatGPT &39.78&7.23&37.13&84.02&\textbf{9.58}&9.49&11.40 &75.40 & 58.96 &50.44\\ 
RAG+Gemini &39.90&9.04&36.97&84.29&\textbf{9.58}&9.65&12.47&\textbf{78.93}&62.91&55.81\\ 
Plain Gemini & 22.60  & 3.22     & 20.85    & 80.81       & 16.38  & 12.72  & 24.18 & 52.44& 53.19       & 44.97    \\
RAG-RLRC  & \textbf{47.91} & \textbf{12.65} & \textbf{44.96} & 84.61 & 10.52 & \textbf{9.11} & 11.73 & 68.61& 61.34 & \textbf{60.40} \\
\midrule
\multicolumn{11}{c}{Average} \\
\midrule
Plain LED  & 46.49 &  13.76  & 42.70  & \textbf{85.35}  & 12.84 & 10.79  & 13.95& 64.20& \textbf{72.02}     & \textbf{67.23}  \\
Plain retrieve+LED &  46.62& 13.38 & 42.68& 85.03 & 13.58 &  10.67 &  13.86 & 59.69 & 65.47 & 64.06 \\
RAG+LED &  46.66& \textbf{13.89}& 43.32 &  84.81 & 13.60 & 10.59 & 13.65 &  60.68 &  65.23 & 64.24\\

RAG+ChatGPT &38.59&7.02&35.55&84.36&\textbf{10.40}&\textbf{9.93}&\textbf{11.95} &73.65 &62.34&54.15\\ 
RAG+Gemini &39.39&8.89&36.04&84.70&10.46&10.06&12.93&\textbf{76.85}&65.66&57.35\\ 
Plain Gemini & 33.63  & 8.29  &30.55   & 83.34 & 16.04  &  12.28  & 21.08 & 57.54 &  63.68 &48.89 \\
RAG-RLRC  & \textbf{47.24} & 13.80 & \textbf{43.38} & 85.22 & 12.70 &10.46 & 14.25 &  58.08& 69.89 &  66.68 \\
\bottomrule
\end{tabular}
\end{center}
\end{table*}

\section{Experimental Settings and Results}
\subsection{Datasets and Evaluation}
This study uses biomedical research articles from the PLOS and eLife datasets, which include both technical abstracts and expert-crafted lay summaries. The PLOS dataset contains 24,773 training and 1,376 validation instances, while the eLife dataset comprises 4,346 training and 241 validation instances \cite{goldsack-etal-2022-making}. We assess summarization quality using predefined metrics: Relevance is gauged by ROUGE Scores (ROUGE-1, ROUGE-2, ROUGE-L) \cite{lin-2004-rouge} and BERTScore \cite{zhang2019bertscore}; Readability by the Flesch-Kincaid Grade Level (FKGL) \cite{Kincaid1975DerivationON}, Dale-Chall Readability Score (DCRS) \cite{dale1948dale}, and Learnable Evaluation Metric for Text Simplification (LENS) \cite{maddela2022lens}; Factuality by Summac \cite{laban2022summac} and AlignScore \cite{zha2023alignscore}.
\subsection{Performance of Baseline Models}
The Plain LED model, serving as our baseline, achieved ROUGE-L scores of 41.33 and 44.07. In contrast, the Plain retrieve+LED model, which integrates external knowledge through the BM25 retriever, slightly improved ROUGE-L scores to 47.02 and 47.21. This indicates that the incorporation of external knowledge slightly enhances the relevance of the summaries.
\subsection{Effect of Neural Re-rankers}
Further improvements were observed with the RAG+LED model, which incorporates a trained neural re-ranker, boosting the ROUGE-L scores to 49.68 and 49.79. This significant increase demonstrates that neural re-rankers are more precise in selecting relevant content, effectively enhancing the accuracy and relevance of the summaries.
\subsection{Effect of Large Language Models}
The RAG+ChatGPT and RAG+Gemini models, utilizing LLMs, achieved high FKGL readability scores of 9.93 and 9.25 respectively, but their ROUGE-L scores were lower at 39.59 and 39.20, indicating that LLMs can sometimes introduce irrelevant information. Similarly, the Plain Gemini model, which relies solely on an LLM, scored only 33.63 in ROUGE-L, demonstrating the challenges LLMs face in producing coherent and accurate summaries without mechanisms for precise content selection.

\subsection{Effect of RLRC}
The RAG+RLRC model, integrating reinforcement learning training strategies, achieved a ROUGE-L score of 47.24. It marked an improvement in factual accuracy, with a Summac score of 78.45 compared to the 73.44 of Plain LED. This highlights the effectiveness of reinforcement learning strategies in optimizing the text's factual alignment.
\section{Related Work}
Automatic summarization in the biomedical domain has been extensively studied \cite{du-etal-2019-extracting, DBLP:journals/corr/abs-2005-01795, goldsack-etal-2023-biolaysumm, devaraj2022evaluating}. The primary challenge in this field is simplifying the content of original articles to make them comprehensible to laypersons. While \citet{rosati2023grasum} supplement source documents to aid in generating more comprehensible summaries, and \citet{devaraj2022evaluating} explore how text simplification impacts summary accuracy, introducing a taxonomy of error types and identifying omissions as a prevalent issue, these approaches often overlook the balance between simplification and factuality.

To enhance summary factuality, researchers incorporate factual knowledge from external sources during model training \cite{9739885}, which has proven effective in improving accuracy. \citet{rosati2023grasum} utilize Wikipedia to enrich summaries with additional knowledge, while \citet{poornash2023aptsumm} employ a trained re-ranker to select pertinent information, enhancing the factuality of summaries. 

\section{Conclusion and Future Work}
The RAG-RLRC-LaySum framework effectively simplifies complex biomedical texts, enhancing readability and factual accuracy for lay audiences. It surpasses traditional models, offering new insights into the pivotal role of knowledge retrieval and readability optimization in scientific summarization. Future work will expand the framework’s knowledge sources and refine how knowledge is utilized, potentially broadening its application across various scientific fields. This will further explore the integration of domain-specific knowledge to improve the precision and relevance of summaries.
\section{Limitations}
While the RAG-RLRC-LaySum framework shows promise, it has several limitations. The reliance on external sources like Wikipedia can introduce biases. The framework's computational complexity is high, making real-time applications challenging. Readability metrics like FKGL and DCRS may not fully capture readability for all audiences. Additionally, the generalizability to other domains beyond biomedical texts is uncertain. Lastly, evaluations based on automated metrics may not fully reflect user experience, highlighting the need for human evaluations. Future work should address these limitations by exploring diverse knowledge sources, optimizing efficiency, refining readability metrics, and conducting human evaluations.
\bibliography{custom}
\appendix
\section{Finetuning models}
\label{sec:Finetuning model}

For training the Longformer Encoder-Decoder (LED) model  \cite{beltagy2020longformer}, we utilized the "allenai/led-base-16384" pre-trained checkpoint available on Huggingface's model hub. Our training setup included a configuration that processes 16,384 input tokens and generates outputs limited to 512 tokens. This training was conducted over the course of a single epoch.

In parallel, we employed BioLinkBERT-base \cite{yasunaga2022linkbert} as the foundational language model for processing eLife and PLOS datasets, leveraging its specialized capabilities in understanding biomedical context.

Then, we designed a neural re-ranker based on the ColBERT \cite{khattab2020colbert} and BGE-v2 \cite{li2023making,chen2024bge} scoring mechanism, which refines the results by evaluating the relevance of retrieved documents. The training for this re-ranker was tailored to accept inputs of up to 512 tokens, and it was fine-tuned to generate models by considering the top 5 most pertinent retrieval results.
Futhermore, we define the Flan-T5-Large from huggingface, we use the model "google/flan-t5-large" as the base model. To make use of the control generation, we use the keywords in the article as the expected output to make sure the relevance.

\subsection{Retrieval Augmented Generation}
We conduct the experiment based on the model Longformer Encoder-Decoder (LED) \cite{beltagy2020longformer} which supports an input token length of 16,384 tokens. 
For the basic fine-tuning method, we find out in both the PLOS and eLife data that the re-ranker result will be a higher result in the Rouge-L and a lower score in the FKGL and DCRS score. In that case, indicate the lower the complexity of the description.

We use ColBERT \cite{khattab2020colbert} and BGE-v2 \cite{li2023making,chen2024bge} as two different types of the neural re-ranker.

\subsection{Reinforcement Learning for Readability Control (RLRC)}

By utilizing various control levels for readability within the model-generated results, we focus on understanding how modifications to the readability scores, particularly the Flesch-Kincaid Grade Level (FKGL), impact the final summaries. The Flan-T5 model \cite{chung2024scaling} serves as the primary backbone for text generation. During the inference phase on testing data, where no ground truth is available for the reward mechanism, keywords are used as proxy indicators to ensure that the generated summaries accurately reflect the expected concepts. 

In our model, we define two key mathematical expressions. The first is the Gaussian probability density function, used to estimate the likelihood of a given value within a normal distribution. The expression for this function is:

\begin{equation}
\begin{split}
&\text{calc\_nd}(value, mean) = \\
&\frac{1}{\sigma \sqrt{2\pi}} \exp\left(-\frac{(value - mean)^2}{2\sigma^2}\right)
\end{split}
\end{equation}
This function is essential for assessing how far a data point deviates from the mean and is widely used in statistical analyses.

The second formula defines our reward function, which combines three different scoring metrics—readability score, BERTScore, and text length score—to comprehensively evaluate the quality of the text. The formula is as follows:
\begin{equation}
\small
\begin{aligned}
&\text{reward} = w_r \cdot \text{normalized\_flesch\_scores} + \\
&\quad w_b \cdot \text{all\_bertscore\_scores} + w_l \cdot \text{length\_scores}
\end{aligned}
\end{equation}

Here, $w_r, w_b$ and $ w_l$ are the weight factors for each scoring metric, adjusting the influence of each score in the overall assessment. By default, we set $w_r=0.5, w_b=0.3, w_l=0.2$.

This weighted approach allows us to tailor the scoring criteria to different types of text analysis tasks, accommodating the multifaceted nature of text data.

\section{Retrieval Design}
\label{sec:Retrieval Design}
For the reranking of retrieved documents, we utilize the pyserini package \cite{lin2021pyserini}. Following the approach outlined by \citet{rosati2023grasum}, we employ $enwiki-paragraphs$ for background knowledge.
We first retrieved 20 candidate paragraphs and then rerank the top 5 results.

\subsection{Neural Re-ranker}
In the provided Table \ref{tab:accuracy_datasets}, the performance trends across the eLife and PLOS datasets reveal that neural re-ranking methods (ColBERT and BGE) consistently outperform the traditional BM25 method. Notably, BGE shows a clear upward trend in accuracy from Top1 through Top20 in both datasets. Similarly, ColBERT's performance also exhibits an upward trajectory, although it remains below BGE, indicating a strong but second-tier efficacy among the tested methods. 
\label{Retrieval Table}
\begin{table}[htbp]
\centering
\caption{Accuracy for Neural Re-ranker.}
\label{tab:accuracy_datasets}
\begin{tabular}{ccccc}
\toprule
Dataset & Method & Top1 & Top5 & Top20 \\
\midrule
\multirow{2}{*}{eLife} & BM25 & 10.32& 42.13 &65.24\\
&ColBERT & 15.38 & 53.85 & 76.92 \\
&BGE & 18.53 & 60.19& 78.52 \\
\midrule
\multirow{2}{*}{PLOS} & BM25& 20.33 & 53.74  & 80.12 \\
& ColBERT& 26.09 & 57.97  & 84.06 \\
&BGE &  29.30 & 59.98  & 88.92 \\
\bottomrule
\end{tabular}
\end{table}
\section{Prompts}
\label{sec:Gemini Prompt}
\begin{table}[htbp]
\caption{One shot prompt for ChatGPT 4 and Gemini 1.0.}
\begin{center}
\begin{tabular}{|l|}
\hline
\textbf{System:} You are a layman rephrase; your goal \\
is to rephrase the input and make it easier to \\
read. For example: 'Diabetes is a condition\\
in which the pancreas cannot produce enough\\ 
insulin to feed the body. This is caused by a \\
protein called proinsulin is an ingredient \\
a group of molecules called cysteine thiols.\\
The rephrased result should be: 'Diabetes is a \\
condition where the pancreas doesn't produce \\
enough insulin to meet the body's needs. This \\
happens because of a protein called proinsulin,\\
which consists of a group of molecules known\\ 
as cysteine thiols.'\\
\hline           
\textbf{Input:} Here is the original text I want you to\\
help me to rephrase: \{first generation\}. Make\\
it easier to read and retain as much of the \\
biomedical phrase as possible and have a similar\\
length as the original text. \\

\hline
\end{tabular}
\label{tab:chatgpt_prompt}
\end{center}
\end{table}

\begin{table}[htbp]
\caption{Prompt used for Gemini for article summarization.}
\begin{center}
\begin{tabular}{|l|}
\hline
I will give you a long article in biomedical \\
publications, you should generate an \\
abstractive summarization of this article in \\
one single paragraph. I will also give you the\\
keyphrases in this article, you should try to \\
include as many keyphrases in your generated \\
summarization as possible. The summarization \\
is with an emphasis on catering to non-expert\\
audiences through the generation\\
of summaries that are more readable, containing \\
more background information and less technical \\
terminology. Keyphrases:\{\}, Article:\{\}.\\
\hline
\end{tabular}
\label{tab:long_dependency_example}
\end{center}
\end{table}

\end{document}